\documentclass[sigconf, authorversion , nonacm]{acmart}
\usepackage{algorithmic}
\usepackage{graphicx}
\usepackage{textcomp}
\usepackage{xcolor}
\usepackage{balance}

\usepackage{amsmath,amsfonts,amsthm}
\usepackage{graphicx}
\usepackage{textcomp}
\usepackage{xcolor}
\usepackage{enumitem}
\usepackage{mathtools}

\usepackage[ruled,vlined,linesnumbered]{algorithm2e}

\usepackage{booktabs} 
\usepackage{subcaption} 

\usepackage{fancyhdr}
\usepackage{tabularx}
\usepackage{multirow}
\usepackage{multicol}
\usepackage{mdwlist}
\usepackage{indentfirst}
\usepackage{physics}
\usepackage{tablefootnote}
\usepackage{footnote}
\usepackage[bottom]{footmisc}
\usepackage{hyperref}
\usepackage{makecell}
\usepackage{subcaption}
\usepackage{xspace}
\usepackage{threeparttable}
\usepackage{diagbox}	
\usepackage{bbding} 
\usepackage[hyphenbreaks]{breakurl}
\xspaceaddexceptions{]\}}

\newcommand{\method}{\textsc{SmartSense}\xspace}

\newcommand{\QTE}{QTE\xspace}

\newcommand*{\eg}{e.g.\@\xspace}

\newtheoremstyle{problemstyle}  
        {3pt}                                               
        {3pt}                                               
        {\normalfont}                               
        {}                                                  
        {\bfseries\itshape}                 
        {\normalfont\bfseries:}         
        {.5em}                                          
        {}                                                  
\theoremstyle{problemstyle}

\copyrightyear{2022}
\acmYear{2022}
\setcopyright{acmcopyright}
\acmConference[CIKM '22] {Proceedings of the 31st ACM International Conference on Information and Knowledge Management}{October 17--21, 2022}{Atlanta, GA, USA.}
\acmBooktitle{Proceedings of the 31st ACM International Conference on Information and Knowledge Management (CIKM '22), October 17--21, 2022, Atlanta, GA, USA}
\acmPrice{15.00}
\acmISBN{978-1-4503-9236-5/22/10}
\acmDOI{10.1145/XXXXXX.XXXXXX}

\begin{document}
\settopmatter{printacmref=false}

\title{Accurate Action Recommendation for Smart Home via Two-Level Encoders and Commonsense Knowledge}

\author{Hyunsik Jeon,\hspace{1em} Jongjin Kim,\hspace{1em} Hoyoung Yoon,\hspace{1em} Jaeri Lee,\hspace{1em} U Kang }
\affiliation{%
  \institution{Seoul National University}
  \city{Seoul}
  \country{South Korea}\\
  \institution{\{jeon185,\hspace{0.2em}j2kim99,\hspace{0.2em}crazy8597,\hspace{0.2em}jlunits2,\hspace{0.2em}ukang\}@snu.ac.kr}
}

\begin{abstract}
How can we accurately recommend actions for users to control their devices at home?
Action recommendation for smart home has attracted increasing attention due to its potential impact on the markets of virtual assistants and Internet of Things (IoT).
However, designing an effective action recommender system for smart home is challenging because it requires handling context correlations, considering both queried contexts and previous histories of users, and dealing with capricious intentions in history.
In this work, we propose \method, an accurate action recommendation method for smart home.
For individual action, \method summarizes its device control and its temporal contexts in a self-attentive manner, to reflect the importance of the correlation between them.
\method then summarizes sequences of users considering queried contexts in a query-attentive manner to extract the query-related patterns from the sequential actions.
\method also transfers the commonsense knowledge from routine data to better handle intentions in action sequences.
As a result, \method addresses all three main challenges of action recommendation for smart home, and achieves the state-of-the-art performance giving up to $9.8\%$ higher mAP@$1$ than the best competitor.

\end{abstract}



\maketitle


\begin{figure}[t]
\centering
\begin{subfigure}[b]{0.42\textwidth}
   \includegraphics[width=1\linewidth]{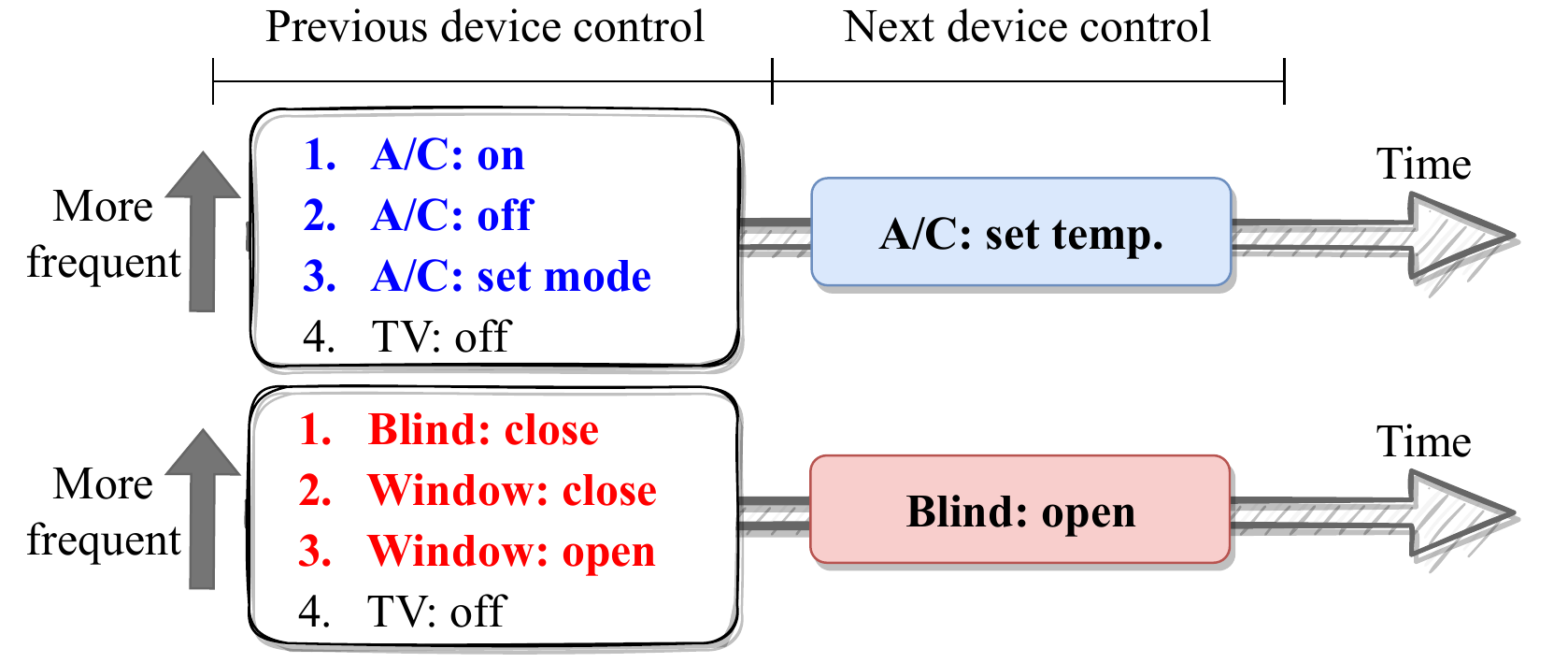}
   \vspace{-0.6cm}
   \caption{Historical dependency of device controls.}
   \label{fig:history-dependent}
\end{subfigure}
\begin{subfigure}[b]{0.42\textwidth}
   \includegraphics[width=1\linewidth]{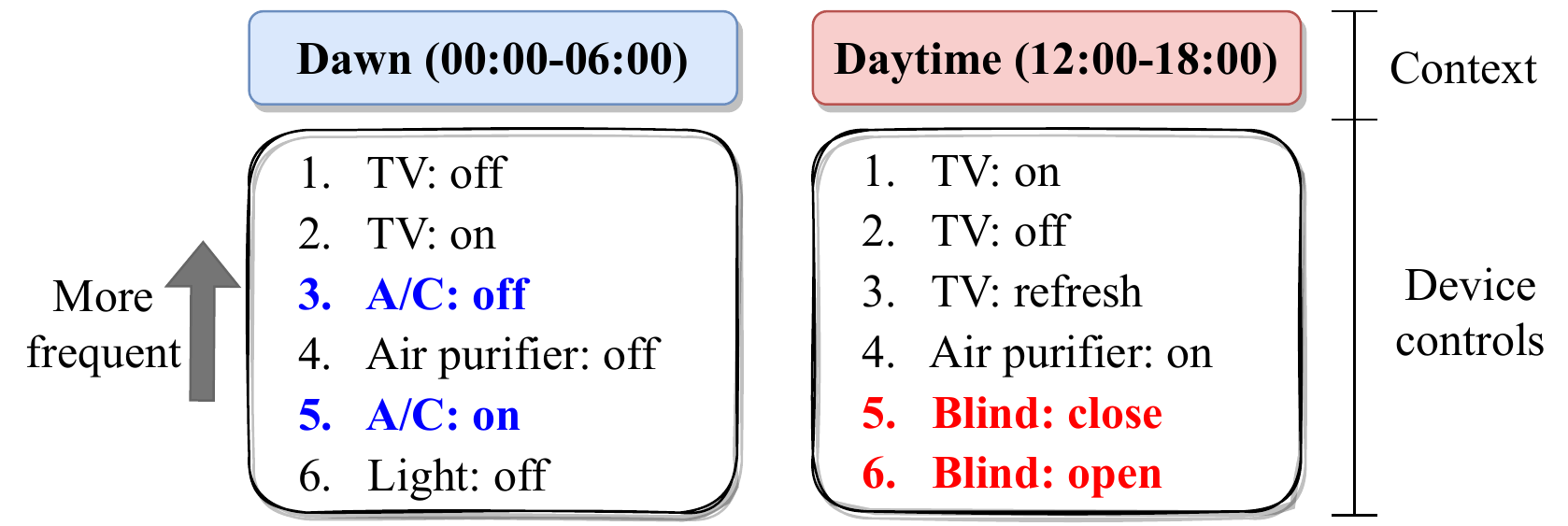}
   \vspace{-0.6cm}
   \caption{Contextual dependency of device controls.}
   \label{fig:context-dependent}
\end{subfigure}
\caption{
	Historical and contextual dependency patterns of SmartThings users.
	(a) Frequent previous device control rankings of two different device controls. The next device controls are affected by previous ones.
	(b) Frequent device control rankings according to two different given hours. Temporal contexts affect users' device controls.
}
\label{fig:two-patterns}
\end{figure}

\begin{figure}[t]
\centering
\includegraphics[width=0.9\linewidth]{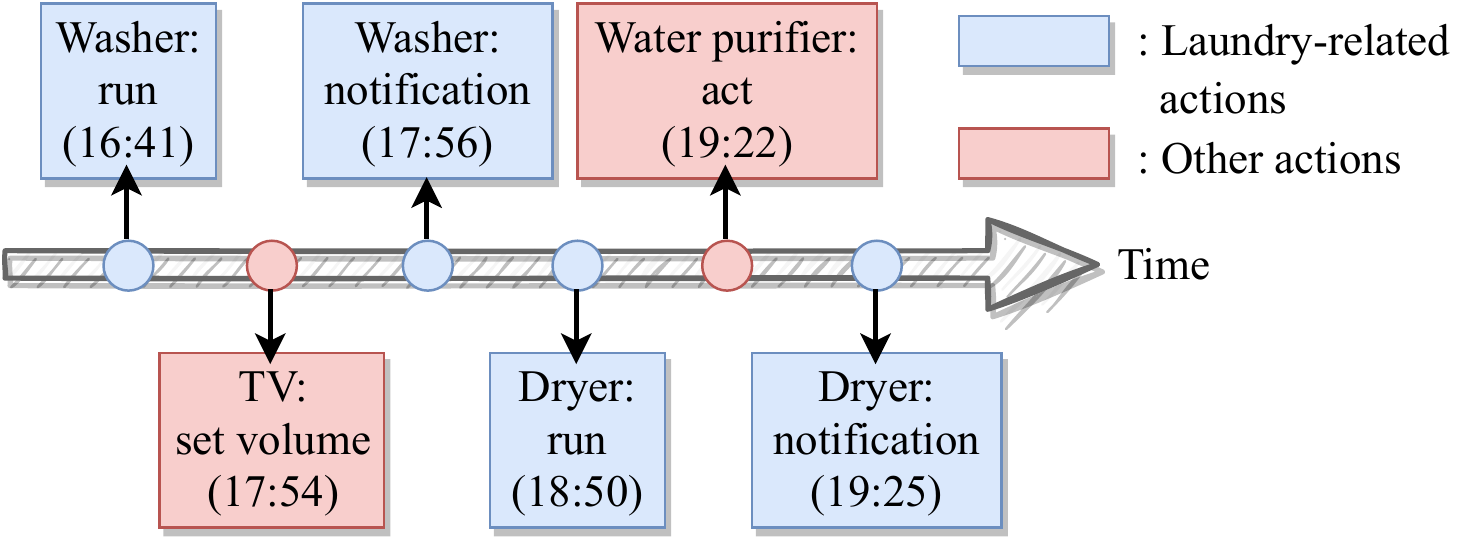}
\caption{
        An example of sequential device controls performed by a \textit{SmartThings} user.
        While the user is controlling the laundry-related devices (washer and dryer), other devices (TV and water purifier) are also controlled since the running times of the laundry-related devices are long.
        A sequence of actions contains capricious intentions.
}
\label{fig:intentions}
\end{figure}

\begin{table}[t]
\caption{\method addresses all three challenges of action recommendation for smart home, while competitors miss one or more of the challenges.}
\label{tab:salesman}
\resizebox{.475\textwidth}{!}{%
\begin{tabular}{l|c|c|c}
\toprule
        \diagbox{Method}{Challenges} & \begin{tabular}{@{}c@{}}Context\\correlations\end{tabular} & \begin{tabular}{@{}c@{}}Context-aware\\personalization\end{tabular} & \begin{tabular}{@{}c@{}}Capricious\\intentions\end{tabular}\\
\midrule
        FMC~\cite{RendelFS10}, TransRec~\cite{HeKM17}, & & \\
        Caser~\cite{tang2018personalized}, SASRec~\cite{kang2018self}, & & \\
        BERT4Rec~\cite{sun2019bert4rec} & & \\
        \midrule
        SIAR~\cite{RakkappanR19} & \CheckmarkBold & &\\
        \midrule
        CA-RNN~\cite{LiuWWLW16} & \CheckmarkBold & \CheckmarkBold &\\
        \midrule
        \textbf{\method{}(proposed)}& \CheckmarkBold & \CheckmarkBold & \CheckmarkBold\\
\bottomrule
\end{tabular}}
\end{table}

\section{Introduction}
\label{sec:introduction}
\textit{How can we accurately recommend actions for users to control their devices at home?}
Action recommendation for smart home has attracted increasing attention in data mining and machine learning communities due to its potential impact on the markets of virtual assistants~\cite{RafailidisM19} and the Internet of Things (IoT)~\cite{ZielonkaWGKPM21}.
The problem is to predict future actions (\eg device controls) of users and recommend the next action. It is of great value, as an accurate recommendation keeps users safe when they forget a critical action (\eg shutting off a gas valve), and reduces the hassles of users when performing a cumbersome action (\eg arming an alarm).

Action recommendation for smart home entails three main challenges.
First, the complicated correlations of multiple contexts such as day of the week, hour affect users' device controls.
For instance, laundry-related controls are highly affected by the correlation of day of the week and hour since people usually do laundry on weekend during the day.
Second, both the sequential pattern of user actions and queried contextual information affects the future user actions.
Figure~\ref{fig:history-dependent} shows the frequency rankings of previously performed device controls in two different cases from SmartThings users, where SmartThings is a smart home system with 62 million active users worldwide.
As shown in the figure, the next actions of people depend on the previous ones.
For instance, people often perform various controls on air conditioners before setting the temperature, while they frequently control blinds and windows before opening the blinds.
Figure~\ref{fig:context-dependent} shows the frequency rankings of device controls depending on two different contexts.
As shown in the figure, the device controls are affected by the current temporal contexts.
For instance, people often control air conditioners before dawn since the proper temperature is an important factor for sleeping, while they frequently control blinds during the day because controlling the amount of sunlight entering the room is important for living.
Figure~\ref{fig:two-patterns} shows that both previous history and the current context are crucial when predicting users' current device controls.
Third, users' action sequences contain capricious intentions. 
For example, if a user performs a series of actions: TV off, blind off, and light off, then we assume that the user wants to sleep.
However, users do not always do their actions with only one intention.
Figure~\ref{fig:intentions} shows an example of an action sequence performed by a SmartThings user.
As shown in the figure, a user controls a TV and a water purifier that are not related to laundry while doing laundry-related actions such as controlling a washer and a dryer.
Capricious intentions lead to a degraded performance of recommendation models since they are trained to treat the actions in a sequence as highly related actions. 

Sequential recommender systems~\cite{RendelFS10, jannach2017recurrent, HeKM17, tang2018personalized, kang2018self, sun2019bert4rec, zhang2021causerec, KooJK20, KooJK21} exploit sequential patterns of user actions to predict the future user actions.
However, they have not addressed any of the three aforementioned main challenges.
Context-aware recommender systems~\cite{LiuWWLW16, RakkappanR19} utilize the contextual information to predict the future user actions.
However, they have not considered both queried contexts and previous histories,
and have not dealt with capricious intentions in the histories.
Thus, there is room for improvement of designing an effective model because the previous works failed to address such challenges.

In this work, we propose \method, a novel approach for action recommendation for smart home.
We design \method with the following main ideas.
First, \method encodes individual action with a device control and its temporal contexts by a context-aware action encoder to consider their complicated correlation. 
Second, \method encodes an action sequence and the current context by a context-attentive sequence encoder to consider both personalization and the contextual information.
Third, \method transfers the knowledge from routines of various users to handle capricious intentions in action sequences.
With these ideas, \method accurately recommends next actions that users would prefer.

Table~\ref{tab:salesman} compares our proposed \method with other methods in various perspectives.
\method is the only method that handles all the three challenges of action recommendation for smart home: context correlations, context-aware personalization, and capricious intentions.

\begin{itemize}[leftmargin=*]
        \item \textbf{Method.}
                We propose \method, an accurate method for action recommendation for smart home.
                \method correlates a device control and its temporal contexts to capture their correlation.
                \method then summarizes a session's history while capturing highly related actions to the current contexts.
                Furthermore, \method deals with capricious intentions by transferring commonsense knowledge from routine data.
        \item \textbf{Experiment.}
                Extensive experiments on real-world datasets show that \method provides state-of-the-art performance with up to $9.8\%$ higher mAP@$1$ in action recommendation for smart home compared to the best competitors (see Table~\ref{table:performance_map}).
        \item \textbf{Case study.}
                We show in case studies that \method successfully recommends actions taking into account the context correlations and context-aware personalization (see Figures~\ref{fig:case_attention} and~\ref{fig:case_context}).

        \item \textbf{Real-world dataset.}
                We open-source the dataset from \textit{SmartThings} which is a worldwide Internet of Things (IoT) platform with 62 million users. 
                This is the first dataset for studying action  recommendation in smart home. 
                We provide sequential device control histories from four countries and device routine data from three territories (see Section~\ref{subsec:experimental_setup}).
                \sloppy
                The datasets are available at \underline{\smash{\url{https://github.com/snudatalab/SmartSense}}}.

\end{itemize}



\section{Related Works}
\label{sec:related_works}

\begin{figure*}[t]
\centering
\includegraphics[width=0.92\linewidth]{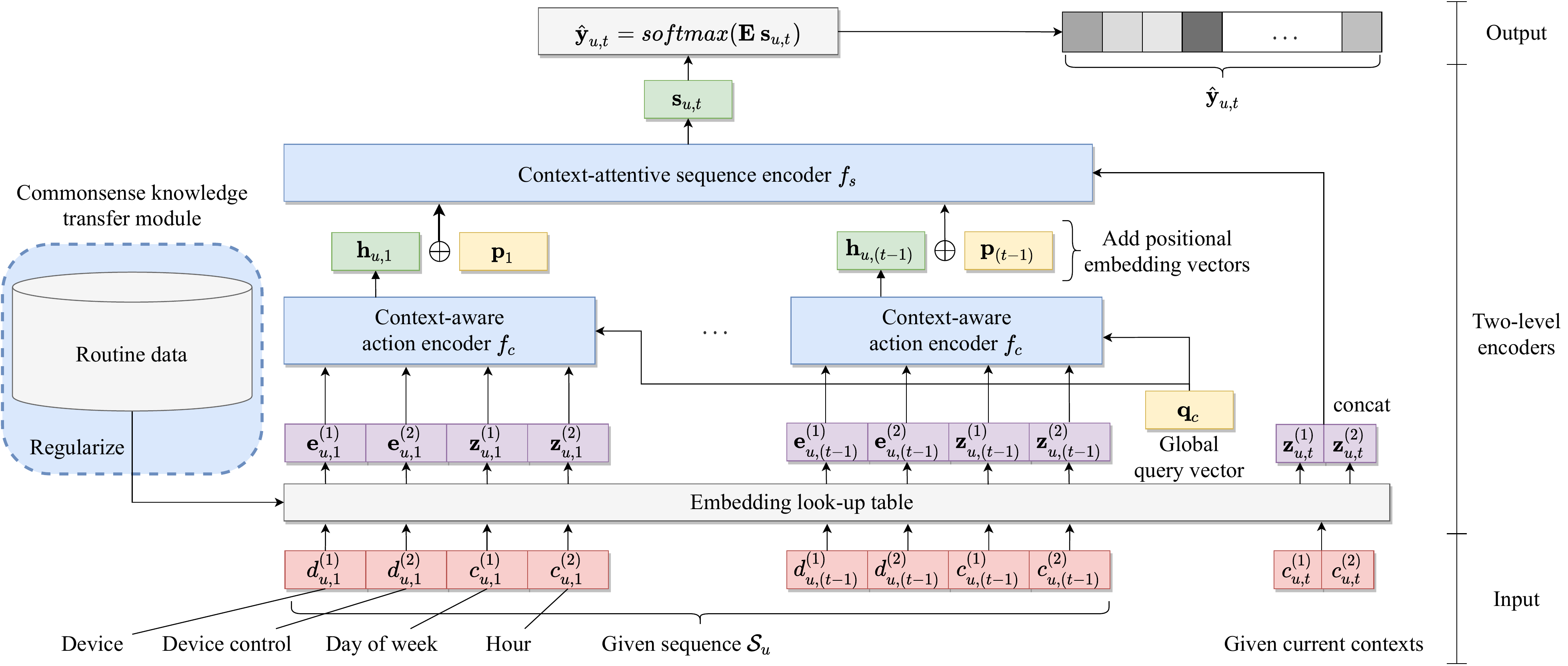}
\caption{
	The overall architecture of \method.
	Given session $u$'th sequence $\mathcal{S}_u$ and the current temporal context $(c^{(1)}_{u,t}, c^{(2)}_{u,t})$, \method predicts the probability $\hat{\mathbf{y}}_{u,t}$ of session $u$'s device control at time $t$.
	\method consists of context-aware action encoder (see Section~\ref{subsec:context_factorization}), context-attentive sequence encoder (see Section~\ref{subsec:sequence_encoder}), and commonsense knowledge transfer module (see Section~\ref{subsec:knowledge_transfer}),
	where each encoder follows the structure of the queried transformer encoder (see Section~\ref{subsec:queried_transformer}).
	Context-aware action encoder interprets an individual action using the global query vector $\mathbf{q}_c$ as a query.
	Context-attentive sequence encoder then summarizes the encoded sequence information using the current context as a query.
	Commonsense knowledge transfer module regularizes device embeddings with routine data.
	Finally, \method estimates the probability of the next device control using the summarized vector and the device control embedding matrix.
	}
\label{fig:architecture}
\end{figure*}

\subsection{Sequential Recommendation}
Given a sequence of user behaviors, sequential recommendation aims to recommend items to users by modeling sequential dependencies of the user behaviors~\cite{WangHWCSO19};
it is different from traditional recommendation systems~\cite{KorenBV09,SalakhutdinovM07,ParkJK17,JeonKK19}, which model interactions in a static way to capture only the general preference of users and items.
On early works, FPMC~\cite{RendelFS10} combines first-order Markov chains (MCs) and Matrix Factorization~\cite{KorenBV09,SalakhutdinovM07} to model both sequential behaviors and general interests of users.
Besides the first-order MCs, higher-order MCs~\cite{HeM16, HeKM17} have been studied to consider previous items in a sequence. 
In recent years, deep neural networks such as Recurrent Neural Networks (RNN)~\cite{HochreiterS97,ChoMGBBSB14}, Convolutional Neural Networks (CNN)~\cite{KrizhevskySH12}, and Transformers~\cite{VaswaniSPUJGKP17} have been adopted in sequential recommendation to address the complicated non-linear patterns in user behaviors.
For instance, GRU4Rec~\cite{HidasiKBT15} used Gated Recurrent Unit (GRU)~\cite{ChoMGBBSB14} to model sequential patterns for session based recommendation.
GRU4Rec$^+$~\cite{HidasiK18} boosted the performance of GRU4Rec for top-k sequential recommendation by improving the loss function.
Caser~\cite{tang2018personalized} employed CNN to capture sequential patterns from both time-axis and feature-axis of sequences.
SASRec~\cite{kang2018self} and BERT4Rec~\cite{sun2019bert4rec} utilized unidirectional Transformers and bidirectional Transformers, respectively, to capture sequential patterns in sequences while considering the importance of correlations between behaviors. 
Advanced techniques such as memory networks~\cite{huang2018improving, ChenXZTCQZ18}, translation learning~\cite{HeKM17}, hierarchical attention mechanisms~\cite{ying2018sequential}, graph neural networks~\cite{wu2019session, ma2020memory, chang2021sequential}, and contrastive learning~\cite{zhang2021causerec} have been adopted to sequential recommendation.
However, such sequential recommender systems fail to achieve high performance in action recommendation for smart home since they do not handle contextual information and capricious intentions in sequences which should not be neglected for an accurate action recommendation.

\subsection{Context-aware Recommendation}
Context-aware recommendation aims to capture user preferences by considering contextual attributes such as time and temperature as well as interaction information between users and items~\cite{baltrunas2011matrix, kulkarni2020context}.
Early works on context-aware recommendation~\cite{xin2019cfm,rendle2011fast} adopted Factorization Machines (FMs)~\cite{Rendle10}, which model all correlations between variables, to capture the significant correlations between contexts.
Recent works leverage deep neural networks and advanced technologies such as attention mechanisms~\cite{xiao2017attentional} to model higher-order interactions and generate more meaningful representations of contexts.
In recent years, context-aware sequential recommendation, which considers sequential patterns as well as contextual information, has been studied.
For instance, CA-RNN~\cite{LiuWWLW16} employs context-specific transition matrix to represent contextual information, and adopts RNN to capture sequential pattern in a history.
Analogously, SIAR~\cite{RakkappanR19} utilizes stacked RNN to consider temporal dynamics of both contexts and actions.
However, such context-aware recommender systems are still not suitable for action recommendation for smart home since they do not simultaneously consider both queried contexts and histories, or unravel the capricious intentions in histories.

\subsection{Transformer}
Transformer~\cite{VaswaniSPUJGKP17} is a deep learning model adopting self-attention mechanism which differentially weights the significance of each part of the input data to handle sequential tasks.
The main advantages of Transformer are that it effectively learns the correlations between the input variables and effectively represents the sequential pattern of data.
Transformer has attracted increasing attention in natural language processing~\cite{DevlinCLT19}, computer vision~\cite{DosovitskiyBKWZUDMHGUH20}, stock prediction~\cite{YooSPK21}, and recommender systems~\cite{kang2018self, sun2019bert4rec} due to its superior performance.
The self-attention mechanism of Transformer enables us to successfully capture significant correlations between a device control and contexts, and effectively represent sequential patterns resulting in accurate action recommendation for smart home.

\section{Proposed Method}
\label{sec:proposed_method}
We define the problem of action recommendation for smart home and propose \method, an accurate method to solve the defined problem.

\subsection{Action Recommendation for Smart Home}
Action recommendation aims to recommend actions to users to help them control their devices at home.
We describe the problem definition of action recommendation for smart home as follows.

\newtheorem{prob}{Problem}
\begin{prob}
	[Action recommendation for smart home]
	For each session $u$, we are given a sequence of history $\mathcal{S}_u= [x_{u, 1},\dots,$\break$ x_{u,(t-1)}]$,
	where $x_{u,i} = (d_{u,i}^{(1)}, d_{u,i}^{(2)}, c_{u,i}^{(1)}, c_{u,i}^{(2)})$ is a quadruplet of indices of $i$'th device, device control, day of week, and hour, respectively.
	Given a previous sequence $\mathcal{S}_u$, and the current temporal contexts $c_{u,t}^{(1)}$ and $c_{u,t}^{(2)}$, the goal is to accurately predict the current device $d_{u,t}^{(1)}$ and its control $d_{u,t}^{(2)}$.
\end{prob}

Note that we aim to predict only the device control $d_{u,t}^{(2)}$, since $d_{u,t}^{(2)}$ also contains the information of the target device $d_{u,t}^{(1)}$.
To address the problem, 
a method should carefully handle device controls and their temporal contexts in the previous sequence, and effectively reflect the current temporal context in the prediction.
Sequential recommendation methods~\cite{RendelFS10,HeKM17,tang2018personalized,kang2018self,sun2019bert4rec} utilize only sequences of device controls without considering any contextual information.
Context-aware recommendation methods~\cite{RakkappanR19,LiuWWLW16} consider temporal contexts in the previous sequence, or deal with the current temporal context.
However, they are still unsatisfactory to action recommendation for smart home since they do not consider capricious intentions in the  previous sequence.

\subsection{Overview}
\label{subsec:overview}
We address the following challenges to achieve a high performance of smart home recommendation.
\begin{itemize}[leftmargin=6mm]
	\item[C1.] \textbf{Considering the correlations of contexts.}
		The complicated correlations between a device control and various contexts such as day of week and hour affect a user's future device control.
		\textit{How can we find meaningful correlations in the device control and the various contexts?}
	\item[C2.] \textbf{Considering both history and the current context.}
		Both a user's past actions and the current context are crucial for predicting the user's current action.
		\textit{How can we personalize the prediction while considering the current contexts?}
	\item[C3.] \textbf{Handling capricious intentions.}
		Capricious intentions make us difficult to learn distant representations for actions with different intentions, which leads to a degraded performance.
		\textit{How can we learn representations of actions such that similar intentions are close to each other and different intentions are distant?}
\end{itemize}

To address the aforementioned challenges, we propose \method with the following main ideas.
\begin{itemize}[leftmargin=6mm]
	\item[I1.] \textbf{Context-aware action encoder (Sections~\ref{subsec:queried_transformer} and \ref{subsec:context_factorization})}.
		We encode a device control and its temporal contexts in a self-attentive manner to capture significant correlations between them.
	\item[I2.] \textbf{Context-attentive sequence encoder (Sections~\ref{subsec:queried_transformer} and~\ref{subsec:sequence_encoder})}.
		We encode a sequence of actions in a self-attentive way to capture the correlations between the actions.
		We then summarize the sequence by a query-attention mechanism to consider the current context in personalization.
	\item[I3.] \textbf{Commonsense knowledge transfer (Section~\ref{subsec:knowledge_transfer})}.
		We transfer commonsense knowledge from routine data.
		Each routine is intentionally defined by a user, as an explicit sequence of actions.
		As a result, the knowledge transfer from the routine data enables us to learn more meaningful representations of actions.
\end{itemize}

Figure~\ref{fig:architecture} shows the overall structure of \method, which consists of context-aware action encoder, context-attentive sequence encoder, and common sense knowledge transfer module.
The context-aware action encoder encodes an action of each time step, and then the sequence encoder predicts the current action of the session based on encoded previous actions and the current context.
The commonsense knowledge transfer module regularizes device embeddings to capture hidden relationships between them.

\begin{figure}[t]
\centering
\includegraphics[width=0.9\linewidth]{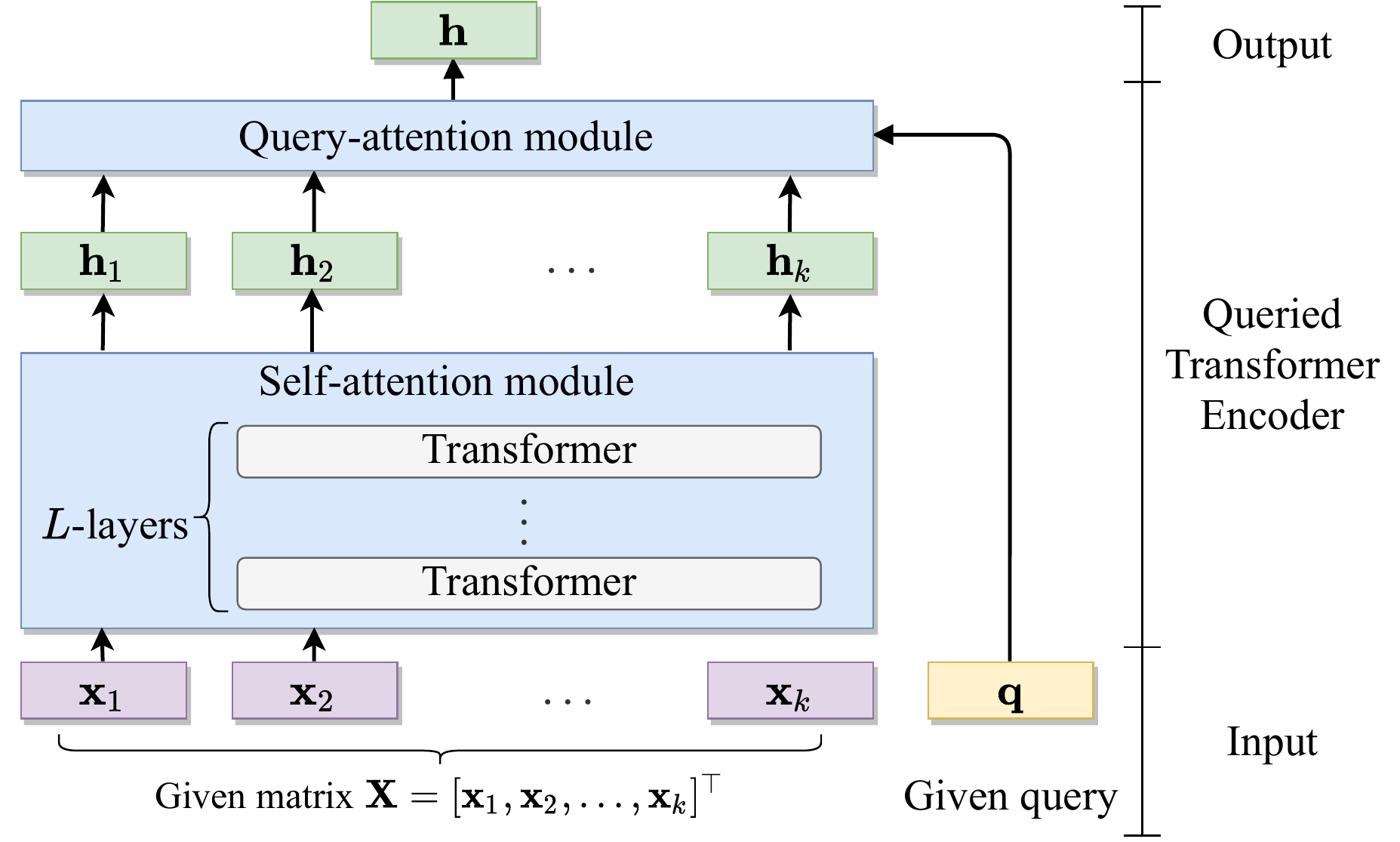}
\caption{
	The queried transformer encoder (\QTE) of \method.
	Given a set of input vectors and a query vector, \QTE summarizes the input vectors into a single output vector while considering the correlations of the input vectors and their relation to the query.}
\label{fig:queried_transformer}
\end{figure}

\subsection{Queried Transformer Encoder}
\label{subsec:queried_transformer}
The aim of action recommendation for smart home is to predict a current device control given a previous history and a current context.
To achieve the goal, we divide the task into two subtasks: 1) encoding individual action, and 2) encoding the sequential history with the current context.
The two subtasks require the two common functionalities as follows.
First, we need to consider the correlations between given variables since multiple variables are intricately related.
Second, we need to consider the significance of each variable because the importance of each variable is different.
Then, how can we design a model to embody the two functionalities?
We propose a queried transformer encoder (\QTE), which is used for both two subtasks: action encoding and sequence encoding.
Our main ideas of \QTE are 1) correlating given variables in a self-attentive manner, and 2) capturing significant variables for a query by a query-attentive mechanism.
\QTE is defined as follows:
\begin{equation}
	\mathbf{h} = f(\mathbf{X}, \mathbf{q}),
\end{equation}
where $\mathbf{h}\in{\mathbb{R}^{d}}$ is the summarized vector, $f(\cdot)$ is \QTE, $\mathbf{X}\in\mathbb{R}^{k\times d}$ is the input matrix, $k$ is the number of input vectors, $d$ is the dimensionality of the input vectors, $\mathbf{q}\in\mathbb{R}^{d'}$ is the query vector, and $d'$ is the dimensionality of the query vector.
$\mathbf{x}_k\in \mathbb{R}^{d}$ denotes the $k$'th row of $\mathbf{X}$ representing the $k$'th input vector.
Figure~\ref{fig:queried_transformer} depicts the structure of \QTE.
\QTE consists of self-attention module and query-attention module which correspond to functionalities of correlating given variables and capturing significant variables, respectively.
We describe the self-attention module and the query-attention module in detail.

\textbf{Self-attention module.}
The goal of self-attention module is to correlate given variables.
We employ transformer encoder~\cite{VaswaniSPUJGKP17} for the self-attention module since it represents all pair-wise correlations between the given variables by learning different query, key, and value weight matrices for each variable.
Given an input matrix $\mathbf{X}$,
we obtain query, key, and value matrices as follows:
\begin{equation}
	\mathbf{Q}=\mathbf{X}\mathbf{W}^Q, \mathbf{K}=\mathbf{X}\mathbf{W}^K, \mathbf{V}=\mathbf{X}\mathbf{W}^V,
\end{equation}
where $\mathbf{Q}, \mathbf{K}, \mathbf{V} \in \mathbb{R}^{k\times d}$ are query, key, and value matrices, respectively;
$\mathbf{W}^Q, \mathbf{W}^K, \mathbf{W}^V \in \mathbb{R}^{d\times d}$ are learnable weight matrices for query, key, and value, respectively.
We compute the transformed matrix as follows:
\begin{equation}
	\bar{\mathbf{X}} = \mathbf{A}\mathbf{V} \quad\text{where}\quad
	\mathbf{A} = \text{softmax}\left(\frac{\mathbf{Q}\mathbf{K}^\top}{\sqrt{d}}\right).
\end{equation}
$\bar{\mathbf{X}}\in \mathbb{R}^{k\times d}$ is the transformed matrix, $\mathbf{A}\in\mathbb{R}^{k\times k}$ is an attention score matrix for all pairs between the variables, and $\text{softmax}(\cdot)$ indicates the row-wise softmax function.
We then adopt a position-wise feed forward network (FNN) and residual connections as follows, to impose nonlinearity in the transformation and enable it to learn an identity function if needed:
\begin{equation}
	\mathbf{H} = \text{Trans}(\mathbf{X})=\mathbf{X} + \bar{\mathbf{X}} + \text{FNN}(\mathbf{X} + \bar{\mathbf{X}}),
\label{eq:transformer}
\end{equation}
where $\mathbf{H}\in \mathbb{R}^{k\times d}$ is the hidden representation matrix of the input variables, $\text{Trans}(\cdot)$ is the transformer, $\text{FNN}(\cdot)$ is a 2-layered position-wise feed forward network with the structure $\mathbb{R}^{d} \rightarrow \mathbb{R}^{4d} \rightarrow \mathbb{R}^{d}$.
We adopt multi-head attention in $\mathbf{Q}$, $\mathbf{K}$, and $\mathbf{V}$ because the multi-head attention shows better performance than a single-head attention in various works such as recommender systems~\cite{sun2019bert4rec}, natural language processing~\cite{DevlinCLT19}, and computer vision~\cite{abs-2101-01169}.
We also adopt dropout and layer normalization after the attention and FNN to improve the generalization performance.
We stack the transformer layer (Equation~\refeq{eq:transformer}) $L$ times to represent the complicated relationships between the input variables.

\textbf{Query-attention module.}
We have the hidden representation matrix $\mathbf{H}\in\mathbb{R}^{k\times d}$ and a query vector $\mathbf{q}\in\mathbb{R}^{d'}$.
We denote $i$'th row vector in $\mathbf{H}$ as $\mathbf{h}_i\in \mathbb{R}^{d}$, and each $\mathbf{h}_i$ corresponds to the hidden representation vector of $\mathbf{x}_i$.
The goal of query-attention module is to summarize the hidden representation matrix $\mathbf{H}$ into a single vector $\mathbf{h}\in\mathbb{R}^{d}$ while capturing significant variables depending on the query vector $\mathbf{q}$.
In other words, we need to give more weight to information relevant to the query since the input variables do not equally contribute to the result.
We propose a query-attention module as follows:
\begin{equation}
\begin{multlined}
	\mathbf{h} = \text{QueryAtt}(\mathbf{H}, \mathbf{q}) = \sum_{i=1}^{k}\alpha_{i}\mathbf{h}_{i},\quad
	\text{where}\quad\\
	\alpha_{i} = \frac{\exp(\beta_{i})}{\sum_{j=1}^{k}\exp(\beta_{j})},\quad
	\beta_{i}=\mathbf{q}^\top \tanh(\mathbf{W}^{H}\mathbf{h}_i+\mathbf{b}^{H}).
\end{multlined}
\end{equation}
$\mathbf{h}\in \mathbb{R}^{d}$ is the summarized vector,
$\mathbf{H}\in \mathbb{R}^{k\times d}$ is the hidden representation matrix,
$\mathbf{q}\in \mathbb{R}^{d'}$ is the query vector,
$\mathbf{h}_i\in \mathbb{R}^{d}$ is $i$'th row vector of $\mathbf{H}$,
$\alpha_i, \beta_i\in\mathbb{R}$ are a normalized and an unnormalized scores for $\mathbf{h}_i$, respectively,
$\mathbf{W}^{H}\in\mathbb{R}^{d'\times d}$ and $\mathbf{b}^{H}\in \mathbb{R}^{d'}$ are learnable parameters, and $\tanh(\cdot)$ is the hyperbolic-tangent function.
We first apply a linear projection using the trainable parameters $\mathbf{W}^H$ and $\mathbf{b}^H$ to $\mathbf{h}_i$, to transform it into the space of the query vector $\mathbf{q}$.
The hyperbolic tangent function makes the transformed vector give element-wise weights to the query vector $\mathbf{q}$.
Finally, we perform the weighted sum of the hidden vectors while considering the importance of each hidden vector $\mathbf{h}_i$.
As a result, we obtain the summarized vector $\mathbf{h}$ while considering the correlations between the input vectors of $\mathbf{X}$ using the self-attention module, and capturing the significant input vectors depending on the query vector $\mathbf{q}$ using the query-attention module.


\subsection{Context-aware Action Encoder}
\label{subsec:context_factorization}

The objective of the context-aware action encoder is to encode the information of an individual action.
To effectively encode each action, the encoder requires two functionalities as follows.
First, it is necessary to correlate a device control and its temporal contexts.
For instance, assume one opens a blind on Monday morning.
We need to correlate opening the blind with morning since the correlation is more important than each of them.
Second, it is required to give more weight to the significant information in an action.
For example, suppose a user turns on an air conditioner on Monday night.
In the action, turning on the air conditioner is more important than the other information because the user is likely to turn off the air conditioner in the future.
Then, how can we consider the two functionalities when encoding an action?

For $i$'th action in session $u$, we have a quadruplet of indices $(d_{u,i}^{(1)}, d_{u,i}^{(2)}, c_{u,i}^{(1)}, c_{u,i}^{(2)})$,
where each of them are indices of device, device control, day of week, and hour, respectively.
We firstly gain each embedding vectors $\mathbf{e}_{u,i}^{(1)},\mathbf{e}_{u,i}^{(2)},\mathbf{z}_{u,i}^{(1)},\mathbf{z}_{u,i}^{(2)}\in \mathbb{R}^{d}$ for the given indices which
correspond to the embedding vectors of $d_{u,i}^{(1)}$, $d_{u,i}^{(2)}$, $c_{u,i}^{(1)}$, and $c_{u,i}^{(2)}$, respectively.
To embody the aforementioned functionalities, we propose to employ the queried transformer encoder (\QTE) for the context-aware action encoder as follows:
\begin{equation}
	\mathbf{h}_{u,i} = f_c(\mathbf{X}_{u,i}, \mathbf{q_c}),
\end{equation}
where $\mathbf{h}_{u,i}\in \mathbb{R}^{d}$ is the hidden representation vector of $i$'th action in session $u$,
$f_c(\cdot)$ is the context-aware action encoder,
$\mathbf{X}_{u,i}\in \mathbb{R}^{4\times d}=[\mathbf{e}_{u,i}^{(1)}, \mathbf{e}_{u,i}^{(2)}, \mathbf{z}_{u,i}^{(1)}, \mathbf{z}_{u,i}^{(2)}]^\top$ is a matrix of stacked vectors,
and $\mathbf{q}_c\in \mathbb{R}^{d}$ is a trainable global query vector.
The context-aware action encoder $f_c$ uses the structure of \QTE.
We use the same query vector $\mathbf{q}_c$ for all actions in the all sessions because we aim to learn global representations for the significance of device controls and temporal contexts.
As a result, for $i$'th action of session $u$, we obtain the encoded vector $\mathbf{h}_{u,i}$ which considers the correlations between the device control and the temporal contexts, and gives more weight to the significant information.


\begin{table*}[t!]
\centering
\caption{Statistics of Log Datasets.}
\label{tab:log_stat}
\begin{tabular}{llrrrrr}
\toprule
Name & Region & Time period (Y-M-D) & \# Sessions & \# Instances & \# Devices & \# Device controls\\
\midrule
KR & Korea & 2021-11-20 $\sim$ 2021-12-20 & 12,992 & 285,409 & 38 & 272\\
US & USA & 2022-02-22 $\sim$ 2022-03-21 & 4,764 & 67,882 & 40 & 268\\
SP & Spain & 2022-02-28 $\sim$ 2022-03-30 & 1,506 & 15,665 & 34 & 234\\
FR & France & 2022-02-27 $\sim$ 2022-03-25 & 388 & 4,423 & 33 & 222\\
\bottomrule
\end{tabular}
\end{table*}

\subsection{Context-attentive Sequence Encoder}
\label{subsec:sequence_encoder}

The objective of the context-attentive sequence encoder is to encode a sequence while considering the sequential patterns and the current context.
It is important to consider the correlations between actions to capture the meaning of the sequence.
For instance, locking the door comes after closing the window if the user wants to go out. 
This example shows that the meaning of sequence changes even if only one action changes.
It is also critical to be aware of the current context to predict the next action.
For example, assume a user has turned off the light.
After that, the user is likely to open the blind to brighten up the room at daytime, while is likely to close the blind to sleep at night.
Then, how can we consider sequential patterns and the current contexts simultaneously when encoding the sequence?

For each session $u$, we have a set of vectors $[\mathbf{h}_{u,1}, ... , \mathbf{h}_{u,(t-1)}]$, where $\mathbf{h}_{u,i}\in \mathbb{R}^d$ is the hidden representation vector of $i$'th action in the session;
note that $\mathbf{h}_{u,1}, ... , \mathbf{h}_{u,(t-1)}$ are from the context-aware action encoder.
To encode the session $u$, we propose to utilize the queried transformer encoder (\QTE) for the context-attentive sequence encoder as follows:
\begin{equation}
	\mathbf{s}_{u,t} = f_s(\mathbf{H}_{u}+\mathbf{P}, \text{concat}(\mathbf{z}_{u,t}^{(1)}, \mathbf{z}_{u,t}^{(2)})),
\end{equation}
where $\mathbf{s}_{u,t}\in \mathbb{R}^{d}$ is the encoded vector for the current time $t$,
$f_s(\cdot)$ is the context-attentive sequence encoder,
$\mathbf{H}_u\in \mathbb{R}^{(t-1)\times d} = [\mathbf{h}_{u,1}, \dots \mathbf{h}_{u,(t-1)}]^\top$ is a matrix of stacked vectors,
$\mathbf{P}\in \mathbb{R}^{(t-1)\times d}$ is a positional embedding matrix,
$\text{concat}(\cdot)$ is the concatenation,
and $\mathbf{z}_{u,t}^{(1)}, \mathbf{z}_{u,t}^{(2)}\in \mathbb{R}^{d}$ correspond to the embedding vectors of the current temporal contexts $c_{u, t}^{(1)}, c_{u, t}^{(2)}$
where
$c_{u, t}^{(1)}$ and $c_{u, t}^{(2)}$ are day of week and hour at the current time $t$, respectively.
The context-attentive sequence encoder $f_s$ uses the structure of \QTE.
The positional embedding matrix $\mathbf{P}$ allows us to identify the position of the input variable, which leads us to learn sequential patterns.
We define the $\mathbf{P}$ as a trainable matrix as in previous works~\cite{VaswaniSPUJGKP17,sun2019bert4rec} for better generalization performance.


After obtaining $\mathbf{s}_{u,t}$, we compute the probabilities of device controls as follows:
\begin{equation}
	\hat{\mathbf{y}}_{u,t} = \text{softmax}(\mathbf{E}\:\mathbf{s}_{u,t}),
\end{equation}
where $\hat{\mathbf{y}}_{u,t}\in \mathbb{R}^{N_d}$ is the predicted probabilities of device controls at the current time $t$ for user $u$,
$\text{softmax}(\cdot)$ is the softmax function,
$\mathbf{E}\in \mathbb{R}^{N_d\times d}$ is the matrix of device controls for the prediction,
and $N_d$ is the number of device controls.

\begin{table}[t!]
\centering
\caption{Statistics of Routine Datasets.}
\label{tab:routine_stat}
\begin{tabular}{llrr}
\toprule
Name & Region & \# Routines & \# Devices\\
\midrule
AP & Asia-Pacific & 17,773 & 36\\
NA & North America & 26,241 & 35\\
EU & Europe & 23,781 & 28\\
\bottomrule
\end{tabular}
\end{table}

\subsection{Commonsense Knowledge Transfer}
\label{subsec:knowledge_transfer}

The aim of commonsense knowledge transfer module is to refine the embedding vectors based on the intention of actions.
As shown in Figure~\ref{fig:intentions}, histories of actions often contain capricious intentions.
This misleads the model to learn false relationship between two unrelated actions which co-occurred in the same sequence but are performed for different purposes.
To solve this problem, we utilize the routine data that include massive routines of multiple users.
Routine data are collections of frequently used device patterns triggered by various contextual backgrounds.
Devices of each routine share the common intention since users submit routines to conveniently execute multiple actions for specific tasks such as doing the laundry or cooling off the room.
Our idea is to adopt transfer learning mechanism, which is widely used in various domain adaptation tasks~\cite{abs-1911-02685,TanSKZYL18,jeon2021unsupervised,lee2021multi,XuK21}, to utilize the knowledge of the routine data.
Specifically, we perform regularization such that similarities between devices in the same routine to be high while those not in the same routine to be low.
Thus, regularization from the routine data makes embedding vectors of related devices sharing the same intention to be closer to each other.

Let $i$'th routine instance $\mathcal{R}_{i}$ be $[d_{1}, d_{2} \dots, d_{r}]$, where $d_{j}$ is $j$'th device of the instance.
Let $\mathbf{e}_{j} \in \mathbb{R}^{d}$ be the embedding vector of device $d_j$.
Note that the commonsense knowledge transfer module shares the device embeddings with the context-aware action encoder and the context-attentive sequence encoder.
We define the regularization loss to minimize as follows:
\begin{equation}
\mathcal{L}_{reg}=-\sum_{i}\sum_{d_{j}\in\mathcal{R}_i}\left(\log(\sigma(\mathbf{e}_{j}^\top\mathbf{e}_{j+1})) +
\sum_{d_k\in p(\mathcal{R}_i)}\log(\sigma(-\mathbf{e}_{j}^\top\mathbf{e}_{k})) \right),
\end{equation}
where $\sigma(\cdot)$ is the sigmoid function and $p(\mathcal{R}_i)$ is the negative samples of $\mathcal{R}_i$.
For each device $d_j$, we randomly select a set of negative devices of size $m$, which are not in routine $\mathcal{R}_i$.

\subsection{Objective Function}
\label{subsec:training}
We train \method to minimize the cross-entropy loss and the regularization loss as follows:
\begin{equation}
	\mathcal{L}(\mathcal{X},\mathbf{Y})=-\frac{1}{n}\sum_u\sum_i{\mathbf{y}_{u,t}(i) \log \hat{\mathbf{y}}_{u,t}(i)} + \mathcal{L}_{reg},
\end{equation}
where $\mathcal{X} \in \mathbb{R}^{n \times l \times 4}$ is an input tensor,
and $\mathbf{Y} \in \mathbb{R}^{n\times N_d}$ is a matrix of the ground-truth label;
$n$ and $l$ are the numbers of sessions and time steps, respectively,
and $N_d$ is the number of device controls.
$\mathbf{y}_{u,t}\in \mathbb{R}^{N_d}$ is the one-hot vector of the ground-truth label for session $u$,
and $\mathbf{y}_{u,t}(i)\in \mathbb{R}$ is the $i$'th element in the vector.
Similarly, $\hat{\mathbf{y}}_{u,t}\in \mathbb{R}$ is the predicted probability for session $u$.
%

%
%
%

\section{Experiments}
\label{sec:experiments}
We perform experiments to answer the following questions:
\begin{itemize}[leftmargin=6mm]
	\item[Q1.] \textbf{Accuracy.} Does \method show higher accuracy in action recommendation for smart home than baselines?
	\item[Q2.] \textbf{Ablation study.} How do main ideas of \method help improving the performance of recommendation?
	\item[Q3.] \textbf{Case study.} Does \method capture meaningful correlations from each action? How do the recommendation results of \method change according to given contexts?
	\item[Q4.] \textbf{Embedding space analysis.} Does \method successfully learn useful embedding vectors of contexts and devices?
\end{itemize}

\begin{table*}
	\centering
	\caption{%
		MAP@$k$ of \method and competitors for smart-home recommendation on four real-world datasets.
		\method outperforms all competitors in all cases, demonstrating its superiority of action recommendation for smart home.
		Bold and underlined values indicate the best and the second-best accuracies, respectively.
	}
	\small
	\begin{tabular}{l|ccc|ccc|ccc|ccc}
		\toprule
		\multicolumn{1}{c|}{} & \multicolumn{12}{c}{\textbf{mAP@$k$}} \\
		\midrule
		\multirow{2}{*}{\textbf{Model}}
			& \multicolumn{3}{c|}{\textbf{Korea}}
			& \multicolumn{3}{c|}{\textbf{USA}}
			& \multicolumn{3}{c|}{\textbf{Spain}}
			& \multicolumn{3}{c}{\textbf{France}}\\
			& $@1$ & $@3$ & $@5$
			& $@1$ & $@3$ & $@5$
			& $@1$ & $@3$ & $@5$
			& $@1$ & $@3$ & $@5$\\
		\midrule
		POP
			& 0.3416 & 0.4918 & 0.5045
			& 0.1886 & 0.3146 & 0.3737
			& 0.4973 & 0.6337 & 0.6455
			& 0.4949 & 0.5955 & 0.6114\\

		FMC~\cite{RendelFS10}
			& 0.5075 & 0.6391 & 0.6569
			& 0.4581 & 0.6082 & 0.6270
			& 0.4102 & 0.5953 & 0.6015
			& 0.4427 & 0.6330 & 0.6477\\
		TransRec~\cite{HeKM17}
			& 0.3854 & 0.5637 & 0.5830
			& 0.3351 & 0.5240 & 0.5426
			& 0.3819 & 0.6149 & 0.6209
			& 0.4255 & 0.6238 & 0.6393\\
		Caser~\cite{tang2018personalized}
			& 0.5676 & 0.7064 & 0.7213
			& 0.5535 & 0.7051 & 0.7177
			& 0.7906 & 0.8548 & 0.8616
			& 0.7706 & 0.8249 & 0.8295\\
		SASRec~\cite{kang2018self}
			& 0.5763 & 0.7064 & 0.7212
			& 0.5657 & 0.7098 & 0.7228
			& \underline{0.7929} & 0.8570 & 0.8630
			& 0.7740 & 0.8286 & 0.8389\\
		BERT4Rec~\cite{sun2019bert4rec}
			& 0.5927 & \underline{0.7253} & \underline{0.7393}
			& 0.5630 & 0.7121 & 0.7254
			& 0.7887 & \underline{0.8610} & \underline{0.8662}
			& \underline{0.7776} & \underline{0.8475} & \underline{0.8507}\\
		CA-RNN~\cite{LiuWWLW16}
			& 0.5703 & 0.6958 & 0.7095
			& 0.4860 & 0.6315 & 0.6459
			& 0.6748 & 0.7253 & 0.7350
			& 0.5141 & 0.5650 & 0.5767\\
		SIAR~\cite{RakkappanR19}
			& \underline{0.5936} & 0.7248 & 0.7381
			& \underline{0.5718} & \underline{0.7163} & \underline{0.7288}
			& 0.7913 & 0.8560 & 0.8628
			& 0.7706 & 0.8258 & 0.8311\\
		\midrule
		\textbf{\method (proposed)}
			& \textbf{0.6515} & \textbf{0.7650} & \textbf{0.7760}
			& \textbf{0.6247} & \textbf{0.7541} & \textbf{0.7639}
			& \textbf{0.8101} & \textbf{0.8707} & \textbf{0.8756}
			& \textbf{0.7944} & \textbf{0.8544} & \textbf{0.8578}\\
		\bottomrule
	\end{tabular}
	\label{table:performance_map}
\end{table*}

\subsection{Experimental Setup}
\label{subsec:experimental_setup}

We introduce our experimental setup: datasets, baselines, evaluation metrics, experimental process, and the hyperparameters.

\textbf{Datasets.}
We use real-world SmartThings datasets collected by Samsung from various regions to evaluate the performance. 
There are four log datasets and three routine datasets.
Log datasets contain histories of device controls executed by users of Bixby.
These are used to generate sequential instances for general sequential recommendation.
As explained in Section 3.6, routine datasets are collections of frequently used device patterns triggered by various contextual backgrounds.
Those instances are submitted by users of SmartThings.
These datasets are used to gain commonsense knowledge in the corresponding log datasets.

Tables \ref{tab:log_stat} and \ref{tab:routine_stat} show the statistics of log datasets and routine datasets, respectively.
In the routine datasets, AP and NA correspond to KR and US in log datasets, respectively.
EU corresponds to SP and FR in log datasets.

\textbf{Baselines.}
We compare \method with existing methods of sequential recommendation and context-aware recommendation.
\begin{itemize}[leftmargin=*]
\item \textbf{POP} recommends device controls based on their popularity.
\item \textbf{FMC ~\cite{RendelFS10}} uses two item embeddings to build an item transition matrix which predicts next device control.
\item \textbf{TransRec~\cite{HeKM17}} represents relationships between consecutive items as a vector operation to perform sequential recommendation.
\item \textbf{Caser~\cite{tang2018personalized}} employs CNN~\cite{KrizhevskySH12} in both time-axis and feature-axis to capture temporal dynamics in a sequential recommendation.
\item \textbf{SASRec~\cite{kang2018self}} uses time-growing directional transformer encoder to consider sequential patterns of user actions.
\item \textbf{BERT4Rec~\cite{sun2019bert4rec}} utilizes BERT architecture~\cite{DevlinCLT19} into sequential recommendation.
\item \textbf{SIAR~\cite{RakkappanR19}} stacks RNN layers of contexts and actions to consider temporal dynamics of both contexts and actions.
\item \textbf{CA-RNN~\cite{LiuWWLW16}} uses context-specific transition matrix in RNN cell to consider context-dependent features in a sequential recommendation.
\end{itemize}
Note that these baselines are not aware of commonsense knowledge of action recommendation for smart home.
Thus, baseline methods use only log datasets during evaluation.

\textbf{Evaluation metrics.}
We evaluate the performance of competing models with two evaluation metrics: hit ratio (HR@$k$) and mean average precision (mAP@$k$).
Both metrics compare the recommendation list of the model with the ground truth value.
For the $k$ size of recommendation list, HR@$k$ treats every item in them equally important, while mAP@$k$ treats higher-ranked items more importantly.
We vary $k$ in $\{1, 3, 5\}$ for all datasets.

\textbf{Experimental process.}
For each session in a log dataset, we create sequential instances with a window of length $10$.
The first nine events of the window are input of the sequential recommendation model.
Each input event is a pair of a temporal context and a device control information.
Temporal context of an event is a combination of a day of week and hour based on the event's timestamp.
The hour in a context is one of the 8 time ranges of 3 hours length: 0-3, 3-6, 6-9, 9-12, 12-15, 15-18, 18-21, and 21-24.
Device control information is composed of a device and its control.
The device control of the last event is the ground truth for the window.
We randomly split sequential instances into a training, a validation, and a test sets in 7:1:2 ratio.
We train the model until the validation accuracy is maximized, and report the test accuracy.

\textbf{Hyperparameters.}
All models are trained with Adam optimizer~\cite{KingmaB14} with learning rate $0.001$ and $l_2$ regularization coefficient $0.00001$.
For fair comparison, we set the size of embedding vectors to 50 and the size of mini batch $1024$ for all models.
For both the context-aware action encoder and the context-attentive sequence encoder, we set the numbers of both transformer layers and the heads as 2.
We set the dropout ratio to 0.1, and the size of negative samples $m=5$ for the commonsense knowledge transfer module.

\begin{table*}
	\centering
	\caption{%
		HR@$k$ of \method and competitors for smart-home recommendation on four real-world datasets.
		\method outperforms all competitors in most cases, demonstrating its superiority of action recommendation for smart home.
		Bold and underlined values indicate the best and the second-best accuracies, respectively.
	}
	\small
	\begin{tabular}{l|ccc|ccc|ccc|ccc}
		\toprule
		\multicolumn{1}{c|}{} & \multicolumn{12}{c}{\textbf{HR@$k$}} \\
		\midrule
		\multirow{2}{*}{\textbf{Model}}
			& \multicolumn{3}{c|}{\textbf{Korea}}
			& \multicolumn{3}{c|}{\textbf{USA}}
			& \multicolumn{3}{c|}{\textbf{Spain}}
			& \multicolumn{3}{c}{\textbf{France}}\\
			& $@1$ & $@3$ & $@5$
			& $@1$ & $@3$ & $@5$
			& $@1$ & $@3$ & $@5$
			& $@1$ & $@3$ & $@5$\\
		\midrule
		POP
			& 0.3416 & 0.6527 & 0.7095
			& 0.1886 & 0.4872 & 0.7493
			& 0.4973 & 0.7916 & 0.8426
			& 0.4949 & 0.7243 & 0.7955\\
		FMC~\cite{RendelFS10}
			& 0.5075 & 0.7921 & 0.8683
			& 0.4581 & 0.7994 & 0.8811
			& 0.4102 & 0.7966 & 0.8226
			& 0.4427 & 0.8529 & 0.9161\\
		TransRec~\cite{HeKM17}
			& 0.3854 & 0.7649 & 0.8478
			& 0.3351 & 0.7596 & 0.8405
			& 0.3819 & 0.8825 & 0.9085
			& 0.4255 & 0.8586 & 0.9247\\
		Caser~\cite{tang2018personalized}
			& 0.5676 & 0.8711 & 0.9345
			& 0.5535 & 0.8886 & 0.9429
			& 0.7906 & 0.9295 & 0.9591
			& 0.7706 & 0.8859 & 0.9073\\
		SASRec~\cite{kang2018self}
			& 0.5763 & 0.8603 & 0.9240
			& 0.5657 & 0.8862 & 0.9420
			& \underline{0.7929} & 0.9320 & \underline{0.9682}
			& 0.7740 & 0.8938 & 0.9377\\
		BERT4Rec~\cite{sun2019bert4rec}
			& 0.5927 & \underline{0.8825} & \underline{0.9424}
			& 0.5630 & \underline{0.8932} & \textbf{0.9502}
			& 0.7887 & \textbf{0.9461} & \textbf{0.9691}
			& \underline{0.7776} & \textbf{0.9303} & \textbf{0.9460}\\
		CA-RNN~\cite{LiuWWLW16}
			& 0.5703 & 0.8428 & 0.9020
			& 0.4860 & 0.8096 & 0.8718
			& 0.6748 & 0.7906 & 0.8324
			& 0.5141 & 0.6294 & 0.6814\\
		SIAR~\cite{RakkappanR19}
			& \underline{0.5936} & 0.8800 & 0.9369
			& \underline{0.5718} & 0.8918 & 0.9453
			& 0.7913 & 0.9314 & 0.9607
			& 0.7706 & 0.8893 & 0.9119\\
		\midrule
		\textbf{\method (proposed)}
			& \textbf{0.6515} & \textbf{0.8983} & \textbf{0.9454}
			& \textbf{0.6247} & \textbf{0.9079} & \underline{0.9495}
			& \textbf{0.8101} & \underline{0.9413} & 0.9623
			& \textbf{0.7944} & \underline{0.9232} & \underline{0.9379}\\
		\bottomrule
	\end{tabular}
	\label{table:performance_hr}
\end{table*}

\subsection{Recommendation Accuracy (Q1)}
We measure the accuracy of \method and baselines in four real world datasets.
Table \ref{table:performance_map} shows the result in terms of mAP.
Note that \method consistently outperforms baselines in all cases.
Table \ref{table:performance_hr} shows the result in terms of HR@k.
Note that \method shows the best performance in most cases, for various $k$.
These results show that \method is an accurate method for action recommendation for smart home.

%
%
%
%
%
%

\subsection{Ablation Study (Q2)}
We verify the effectiveness of our three main ideas, context-aware action encoder, context-attentive sequence encoder, and commonsense knowledge transfer.
We compare \method with \method-$Act$, \method-$Seq$, \method-$Reg$, and \method-$All$.
\method-$Act$ is a \method without a context-aware action encoder.
This model encodes each pair of a temporal context and a device control as the mean value of corresponding embeddings.
Thus, the model is not aware of correlations of temporal context and device control.
\method-$Seq$ is a \method without a context-attentive sequence encoder.
The sequence encoder of this model summarizes the output of transformer layers as a mean value instead of a weighted sum.
This is equivalent to the context-attentive sequence encoder where the given query is a zero vector instead of the current temporal context embedding.
In this way, the model is not aware of the current temporal context during prediction.
\method-$Reg$ is a \method without the regularization from commonsense knowledge transfer module.
This model cannot access information gathered from routine datasets.
\method-$All$ is a \method without any of the main ideas.

Table \ref{tab:ablation} shows that \method is the most accurate model among competitors,
while \method-$All$ shows the worst accuracy.
In summary, all three main ideas are helpful to improve performance of action recommendation for smart home.

\begin{table}
	\setlength\tabcolsep{3pt}
	\centering
	\caption{%
		Ablation study of \method.
		We report the performances by MAP@$k$.
        Note that \method without at least one of the three main ideas decreases the performance,
        while \method with all the ideas shows the best performance.
	}

	\small
	\begin{tabular}{l|ccc|ccc}
		\toprule
		\multirow{2}{*}{\textbf{Model}}
			& \multicolumn{3}{c|}{\textbf{Korea}}
			& \multicolumn{3}{c}{\textbf{USA}}\\
			& $@1$ & $@3$ & $@5$
			& $@1$ & $@3$ & $@5$\\
		\midrule
		\method-$Act$
			& 0.5925 & 0.7256 & 0.7389
			& 0.5802 & 0.7228 & 0.7350\\
		\method-$Seq$
			& 0.6484 & 0.7631 & 0.7743
			& 0.6194 & 0.7489 & 0.7592\\
		\method-$Reg$
			& 0.6461 & 0.7608 & 0.7721
			& 0.6189 & 0.7497 & 0.7600\\
		\method-$All$
			& 0.5941 & 0.7265 & 0.7396
			& 0.5752 & 0.7198 & 0.7321\\
		\midrule
		\textbf{\method}
			& \textbf{0.6515} & \textbf{0.7650} & \textbf{0.7760}
			& \textbf{0.6247} & \textbf{0.7541} & \textbf{0.7639}\\
		\bottomrule
	\end{tabular}
	\label{tab:ablation}
\end{table}

\begin{figure}
	\centering
	\includegraphics[width=0.9\linewidth]{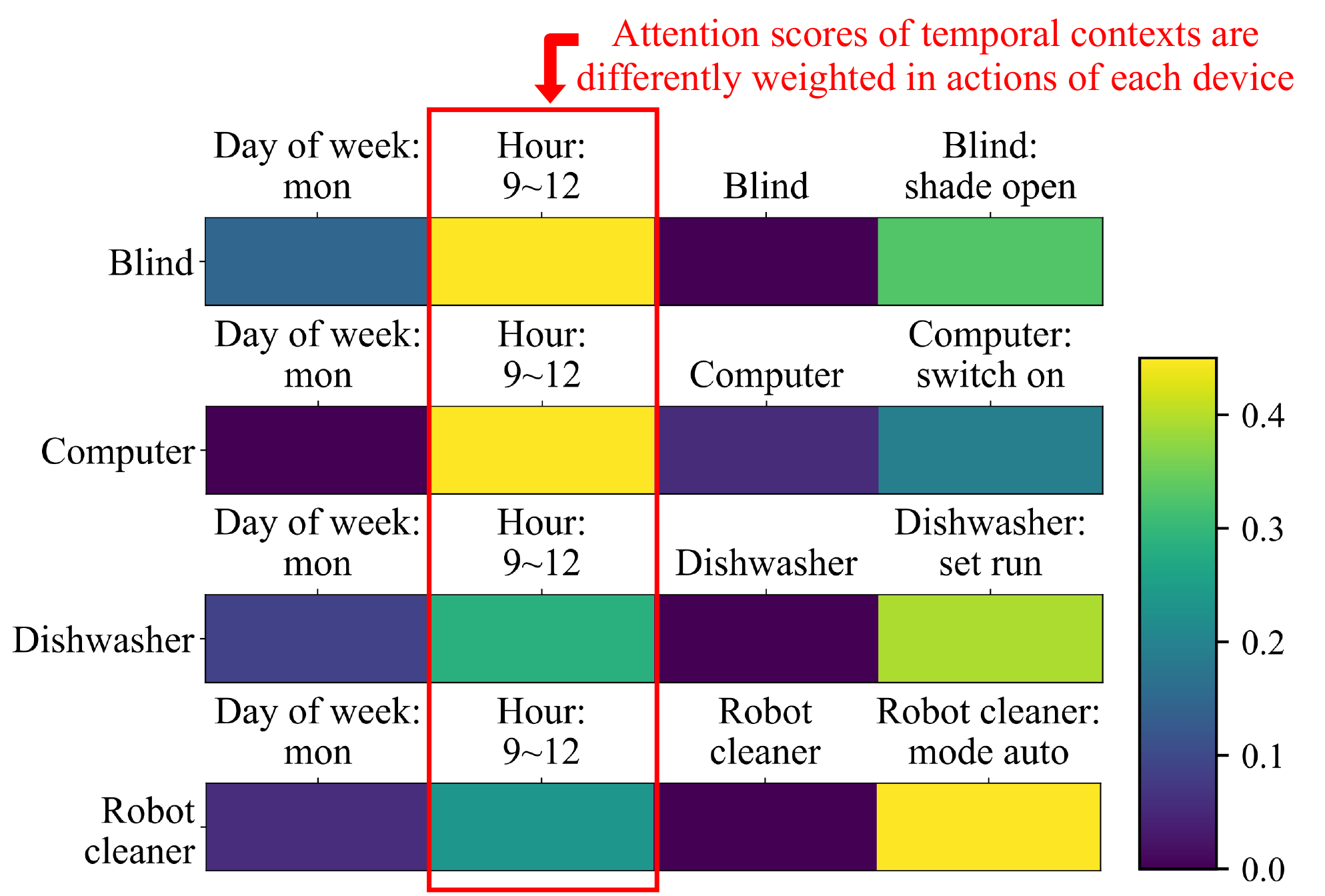}
	\caption{
		Attention scores in the context-aware action encoder.
		The attention scores between hour and time-sensitive devices such as a blind or a computer are high,
		while the attention scores between hour and time-insensitive devices such as a dishwasher or a robot cleaner are low.
		This shows that the encoder successfully captures the relationship of a temporal context and a device.
	}
	\label{fig:case_attention}
\end{figure}

\vspace{-0.3cm}
\subsection{Case Study (Q3)}
We analyze cases to observe how \method deals with context information.

First, we observe the attention scores of the context-aware action encoder to find out how \method reacts to the given pair of temporal context and a device control.
Figure \ref{fig:case_attention} visualizes the attention scores in the last layer of the encoder when the given temporal context is Monday with the hour of 9 to 12.
Each row of the figure corresponds to a device control at a specific temporal context;
the colors show how much attention the encoder gives to day of the week, hour, device, and its control information.
Note that on the first row, the attention score between hour and blind is relatively high since the degree of control for incoming sunlight depends on the time.
The attention score between a computer and hour is also high, because computers are turned on and off according to work hours.
However, dishwasher and robot cleaner do not have strong correlations with hour.
Both dishwasher and robot cleaner can be controlled when users are not paying attention to the device, so they can be controlled at any time.
In summary, \method successfully captures the important correlation from given inputs with a context-aware action encoder.

\begin{figure}
	\centering
	\includegraphics[width=0.9\linewidth]{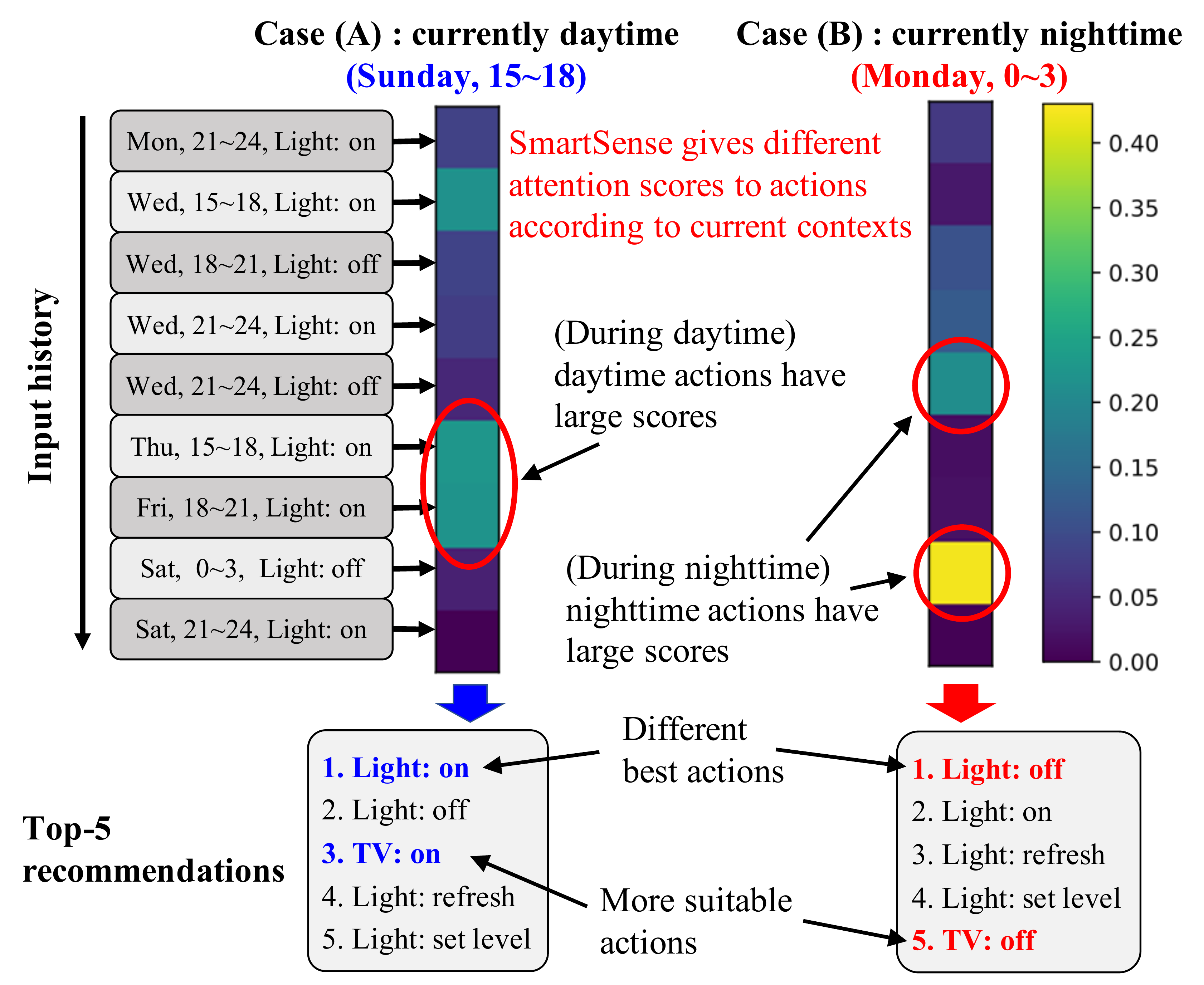}
	\caption{
		Top-5 recommendation lists and attention scores for each action depending on the current temporal context.
			In each case, attention scores of actions with the context similar to the current context are high,
			and \method recommends the device controls of the most attentive actions the most.
			In summary, \method considers the current context to recommend proper device controls for a user.
	}
	\label{fig:case_context}
\end{figure}

Second, we observe how the current temporal context affects recommendation results.
Figure \ref{fig:case_context} shows top-5 recommendation results and attention scores of each action in a history depending on the current temporal context.
The attention scores of the sixth and the seventh actions are high in case (A), while attention scores of the fifth and the eighth actions are high in case (B).
They are actions with the most similar contexts compared to the current context in each case.
\method recommends turning on a light the most in case (A) but it recommends turning off a light the most in case (B),
which are device controls of the most attentive actions.
This shows that \method dynamically collects important information from the input sequence depending on the current temporal context.
Moreover, \method recommends turning on a television in case (A) but it recommends turning off a television in case (B).
In case (A), it is appropriate to recommend turning on the television because a user would be active since the given context is daytime.
However, the given context is nighttime in case (B) so it is reasonable to recommend turning off the television because a user would like to sleep.
This shows that \method understands not only the difference between contexts but also the hidden meanings of each context.
In summary, \method is aware of the current temporal context while prediction, so it flexibly generates an accurate recommendation for the given situation.

\vspace{-0.5cm}
\subsection{Embedding Space Analysis (Q4)}

We observe the embeddings of devices and temporal contexts to analyze how they reflects the real world.

\begin{figure}
	\centering
	\includegraphics[width=0.9\linewidth]{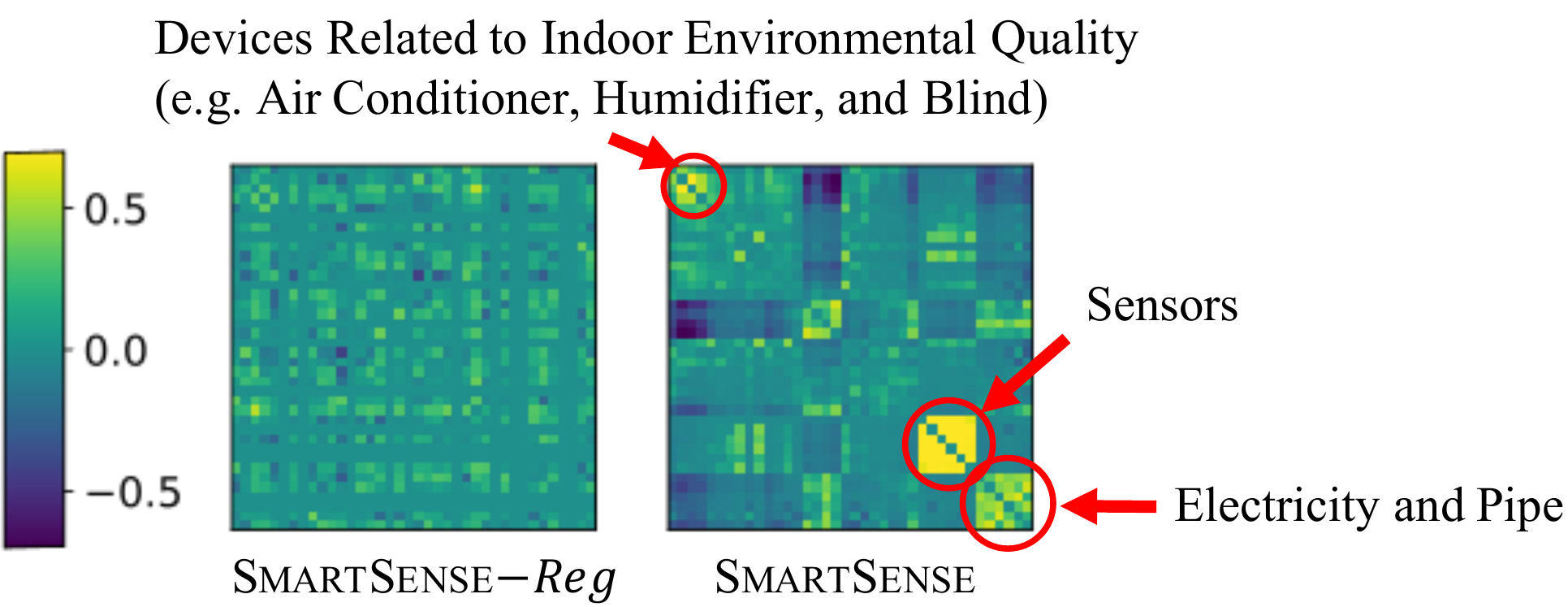}
	\caption{
		Matrix of cosine similarities between different device embedding vectors in \method-$Reg$ (left) and \method (right). 
		Device embedding vectors of similar devices are closer to each other in \method compared to \method-$Reg$, thanks to the common sense knowledge transfer from routine data.
	}
	\label{fig:case_routine}
\end{figure}

First, we observe the device embedding space to verify the impact of commonsense knowledge transfer in the representation learning.
Figure \ref{fig:case_routine} shows that the cosine similarities between embedding vectors of devices in \method-$Reg$ (\method without the commonsense knowledge transfer regularization), and \method.
The standard deviation of cosine similarities in \method is 0.21 while that of \method-$Reg$ is 0.11,
which shows that device embedding vectors in \method-$Reg$ have more complex patterns as the regularizations are applied.
Furthermore, embedding vectors of similar devices in \method such as sensors get closer to each other in \method compared to \method-$Reg$.
This shows that routine data are useful to find close relationships between devices due to its task oriented composition.

\begin{figure}
	\centering
	\includegraphics[width=0.83\linewidth]{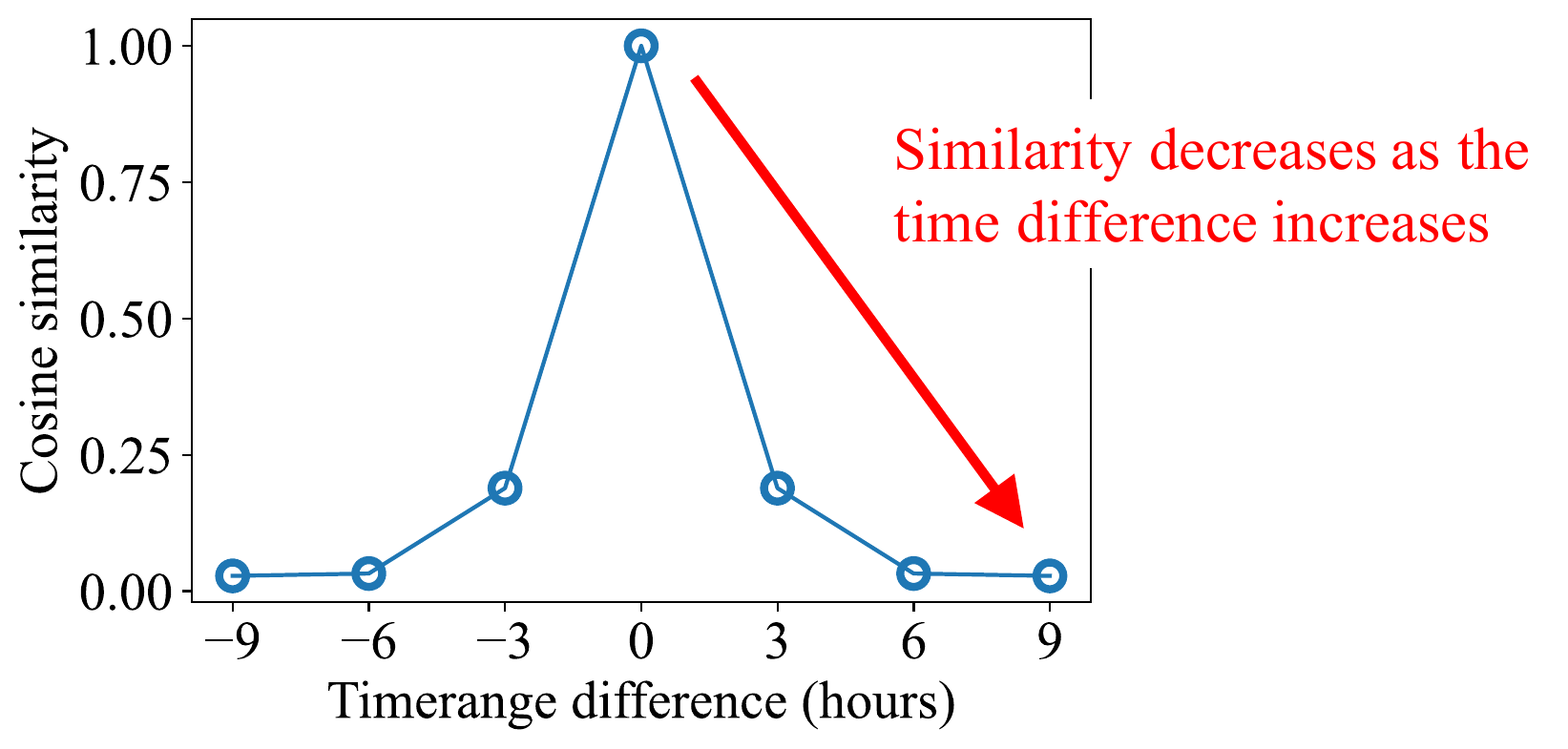}
	\caption{
		Average cosine similarity between two hour embedding vectors depending on their time difference.
		Embeddings of close hours have similar values.
		This shows that embeddings of temporal contexts successfully represent the real world characteristics of corresponding entities.
	}
	\label{fig:case_embedding}
\end{figure}

Second, we visualize the hour embeddings to see whether \method successfully captures the characteristic of temporal contexts.
Figure \ref{fig:case_embedding} shows that embeddings of close hours are more similar to each other compared to farther hours.
This shows \method extracts the crucial contextual information from temporal contexts.

\section{Conclusion}
\label{sec:conclusion}
In this paper, we propose \method, an accurate action recommendation method for smart home.
To reflect the importance of correlations, \method introduces context-aware action encoder which captures significant correlations between the device control and the temporal context.
The context-attentive sequence encoder of \method summarizes users' sequential pattern and queried contextual information to consider both personalization and the current context.
Commonsense knowledge transfer in \method enables the model to successfully consider user intentions.
As a result, \method shows the state-of-the-art accuracy giving up to 9.8\% higher mAP@1 in action recommendation for smart home than the best competitor in extensive experiments.
Through case studies, we also show that \method successfully captures the correlations among actions and contexts.

\begin{acks}
    This work was supported by Samsung Electronics. 
    The Institute of Engineering Research at Seoul National University provided research facilities for this work. The ICT at Seoul National University provides research facilities for this study. 
	We thank Inchul Hwang, Sanghee Kim, and Hyunju Seo for their help and insightful discussion.
	U Kang is the corresponding author.
\end{acks}

\newpage

\bibliographystyle{ACM-Reference-Format}
\balance
\bibliography{paper}

\end{document}